\documentclass[runningheads]{llncs}
\usepackage[T1]{fontenc}
\usepackage{graphicx}
\usepackage{orcidlink}

\usepackage{multirow}

\usepackage{appendix}
\usepackage{float}
\usepackage{longtable}
\usepackage{marvosym}
\usepackage{booktabs}

\usepackage{tabularray}
\usepackage{array}
\newcolumntype{P}[1]{>{\centering\arraybackslash}p{#1}}
\newcolumntype{M}[1]{>{\centering\arraybackslash}m{#1}}

\begin{document}
\title{Understanding Real-World AI Planning Domains: A Conceptual Framework}

%
%
\author{Ebaa Alnazer\,\textsuperscript{\Letter} \orcidlink{0000-0001-6951-4227} \and
Ilche Georgievski\, \orcidlink{0000-0001-6745-0063}}
\authorrunning{E. Alnazer and I. Georgievski}
%
\institute{Service Computing Department, IAAS, University of Stuttgart, Germany \\
Universitätsstrasse 38, 70569 Stuttgart, Germany\\ 
\email{\{ebaa.alnazer,ilche.georgievski\}@iaas.uni-stuttgart.de}}

\maketitle              
\begin{abstract}
Planning is a pivotal ability of any intelligent system being developed for real-world applications. AI planning is concerned with researching and developing planning systems that automatically compute plans that satisfy some user objective. Identifying and understanding the relevant and realistic aspects that characterise real-world application domains are crucial to the development of AI planning systems. This provides guidance to knowledge engineers and software engineers in the process of designing, identifying, and categorising resources required for the development process. To the best of our knowledge, such support does not exist. We address this research gap by developing a conceptual framework that identifies and categorises the aspects of real-world planning domains in varying levels of granularity. Our framework provides not only a common terminology but also a comprehensive overview of a broad range of planning aspects exemplified using the domain of sustainable buildings as a prominent application domain of AI planning. The framework has the potential to impact the design, development, and applicability of AI planning systems in real-world application domains.

\end{abstract}
\section{Introduction}

Artificial Intelligence (AI) planning is the process of finding and organising a course of action to achieve some designated goals~\cite{ghallab2004automated}. The field of AI planning has matured to such a degree that it is increasingly used to solve planning problems in real application domains, such as autonomous driving~\cite{alnazer2022role}, intelligent buildings~\cite{georgievski2017planning}, robotics~\cite{weser2010htn}, and cloud computing~\cite{georgievski2017cloud}. In these application domains, various types of AI planning techniques have been used that differ in their assumptions and applicability for solving different planning problems. All the techniques, however, require gathering and formulating adequate and relevant knowledge of the application domain. This is in fact one of the phases needed to design and develop AI planning systems. Other phases include requirement analysis, selecting a suitable planning type, designing the planning systems, etc.~\cite{georgievski2023:dlaip}.

A crucial process that should precede the development of an operational and successful planning system is identifying and understanding relevant and realistic planning aspects that capture the complexity and characteristics of the application domain without making any simplified assumptions. This process is essential as it needs to guide all the phases of the development process of a planning system that can be utilised in actual settings. However, this process is difficult because AI planning covers a broad range of aspects considering the physics, functions, and qualities of the application domain. Despite all this, there are currently no mechanisms that can support software engineers and knowledge engineers in this process.

Our aim is, therefore, to conceptualise the \textit{realism} of planning domains. To this end, we develop a top-down approach in which we explore and analyse the characteristics of planning domains in the existing literature. The outcome is a 
\textit{conceptual framework} that contains realistic aspects of planning domains and corresponding categories with varying levels of granularity. The framework would form the basis of the design, identification, and categorisation of elements (e.g., planning domain models, planning software components or services, provenance data) for real-world planning applications. In particular, the benefits of the realistic-aspects framework can be summarised as follows.

\begin{itemize}
    \item It helps to advance towards a common and inclusive notion of the realism of AI planning domain models. 
    \item It serves as a basis for characterising planning problems based on their requirements. This offers useful guidelines for planning and software engineers through all the phases of planning system development on which methods and tools can reflect the requirements.
    \item It can drive the development of AI planning techniques and tools to address real-world planning problems' aspects.
    \item It provides means for comparing different AI planning systems based on their support of the real-domain aspects.
    \item It lays the groundwork for other AI planning research on the topics of improving the applicability of AI planning in real-world applications and guiding planning engineers in the development of planning systems.
    \item It highlights some aspects simplified in the existing literature on AI planning.
\end{itemize}

The rest of the paper is organised as follows. Section~\ref{sec:preliminaries} provides the necessary fundamentals and the problem statement. Section~\ref{sec:methodology} presents our methodology to develop the conceptual framework. Section~\ref{sec:realistic-aspects} introduces our conceptual framework by focusing on the categories at the highest level of granularity. Finally, Section~\ref{sec:conclusions} concludes the paper with a discussion of our findings and future work.
\section{Fundamentals and Problem Statement}\label{sec:preliminaries}
We provide a brief introduction to AI planning followed by discussions on planning domain knowledge and designing AI planning systems, where we highlight the importance of having and understanding realistic aspects of planning domains. This serves as our basis to then state the problem our work focuses on.

\subsection{Artificial Intelligence (AI) Planning}\label{sec:aip}
Artificial Intelligence (AI) planning is a subfield of AI that focuses on researching and developing planning systems that aim to find, organise and execute a course of action, i.e., a plan, in order to achieve some designated goal~\cite{ghallab2004automated}. Depending on how complex and realistic the application domain is, one can employ various types of planning. The most basic but widely used type of planning is classical planning. It is based on the concept of actions and makes restrictive assumptions about how the environment of the application domain looks like. In particular, the environment is fully controllable, observable, deterministic, and static (no exogenous events), without temporal properties (actions are instantaneous), and plans are linearly ordered sequences of actions. Other planning types aim at relaxing some of these assumptions. Examples include probabilistic planning, which allows actions to have probabilistic effects, and temporal planning, which allows actions to have durations and considers the temporal interaction between them. Hierarchical Task Network (HTN) planning is another type of planning but one that breaks with the tradition of classical planning by introducing a hierarchy over actions with the help of tasks that can be refined into smaller subtasks using so-called decomposition methods that represent specific knowledge from the application domain~\cite{georgievski2015htn}.

\subsection{Planning Domain Knowledge}
AI planning is a knowledge-based technique, meaning, to compute plans, AI planning systems require relevant and adequate knowledge about the application domain. The knowledge consists of a \textit{planning domain model} and an associated \textit{problem instance}. A planning domain model is a formal representation of the domain knowledge, which is an abstract and conceptual description of the application domain. A problem instance is a specification of a particular planning scenario to be solved within this domain.

In classical planning, a planning domain model formalises the domain knowledge in terms of domain objects with their relations and properties, and \textit{actions} that can change the state of the environment. A problem instance is specified via an initial state and set of goal states that need to be reached. The planning domain models and problem instances used by other types of planning support more constructs, thus enabling the expression of more complex and realistic domain knowledge. For example, in HTN planning, the planning domain model is formalised in terms of actions, \textit{compound tasks}, and \textit{methods}. Actions are defined the same as in classical planning. Compound tasks are more complex tasks than actions and need to be refined into smaller tasks utilising methods. Methods enable encoding of how compound tasks can be achieved by achieving smaller tasks through the means of specific domain knowledge.

\subsection{Designing AI Planning Systems}

The design and development of a typical AI planning system can go through various phases~\cite{georgievski2023:dlaip}. In the first phase, relevant requirements should be analysed. The requirements can be functional, non-functional, user-related, and domain-oriented. Having relevant and well-defined requirements is of utmost importance as it affects the suitability of the intended planning system to address the \textit{real-world aspects} of the application domain. So, this phase is crucial as it provides the ingredients necessary to select a suitable planning type, design a planning domain model, design the system architecture, and define relevant provenance data. The main concern of the second phase is the selection of a suitable planning type. The selection depends on the assumptions about how \textit{realistic} the environment is (see Section~\ref{sec:aip}). In the third phase, the requirements are used to formulate planning domain knowledge out of which a planning domain model is created. The proper execution of this phase in terms of detailed knowledge encoding and management is essential as the lack of relevant or ill-described knowledge can lead to planning domain models that do not reflect the \textit{intended} aspects of the corresponding application domain~\cite{vaquero2013itsimple}. This can lead, eventually, to unsatisfactory plans that cannot be executed in real settings~\cite{vaquero2010improving}. In the fourth phase, the planning system is designed, where the choice of relevant software-engineering principles, design approaches and patterns, and other specific techniques is dictated by the type and nature of the output of the previous three phases. For example, if we want to develop a planning system for any planning domain, e.g., sustainable buildings, that computes plans, schedules, and executes the plan actions in real-time, then we need to \textit{adequately} structure and connect \textit{relevant} planning components.

\subsection{Problem Statement}
Understanding and gathering the relevant aspects characterising real-world planning problems is crucial to the design, development, and applicability of AI planning systems. In particular, this should precede all development phases of AI planning systems since these relevant aspects should be considered and reflected during the execution of each development phase. The way these aspects are reflected depends on the particularities of each phase. For example, in the third phase, where planning engineers create a domain model out of the planning knowledge they acquired, it is necessary to see how the different aspects are reflected in the planning constructs used in the model. Similarly, in phase 5, when selecting suitable AI planning tools, planning engineers should have adequate knowledge about the aspects that the planning system should support to be able to select the tools that support these aspects. There exist some works that focus on providing means to support the design of planning domain knowledge. These works assume that relevant requirements and specifications of relevant domain knowledge are already given by stakeholders or domain experts. The aim of knowledge engineers is to satisfy these requirements and specifications in their design of planning domain knowledge. Usually, the process of designing this knowledge is done in an ad-hoc manner, and the quality of the resulting planning domain models depends mainly on the skills of the knowledge engineers and, if available, the tools they use~\cite{mccluskey2017engineering,vallati2021quality}. In this context, a quality framework is suggested aiming at developing systematic processes that support a more comprehensive notion of planning domain quality~\cite{vallati2021quality}. Some other studies focus on conceptualising planning functionalities as distinct software components so that they can be directly and flexibly used to address the intended requirements of application domains (e.g.,~\cite{georgievski2022:aipas}).

However, to the best of our knowledge, there is currently no support for software engineers and knowledge engineers in the process of identifying relevant and realistic aspects of real-world planning domains necessary for the development of essential planning elements (i.e., requirements, planning types, planning domain models, planning system design). Our work is positioned within this research gap and aims at answering the following research question: \textit{What are the realistic aspects that should be considered in the process of developing AI planning systems for real-world domains and how those aspects can be meaningfully organised?}
\section{Approach}\label{sec:methodology}
To address our research question, we perform top-down exploratory research on existing literature to find relevant information related to the realism of planning domains and create the conceptual framework of realistic aspects for planning domains. We illustrate the elements of the conceptual framework using the domain of \textit{Sustainable Buildings}, which is one prominent example of real-world application domains.

\subsection{Methodology}
\begin{figure}[t!]
	\centering
        \includegraphics[width=0.8\textwidth]{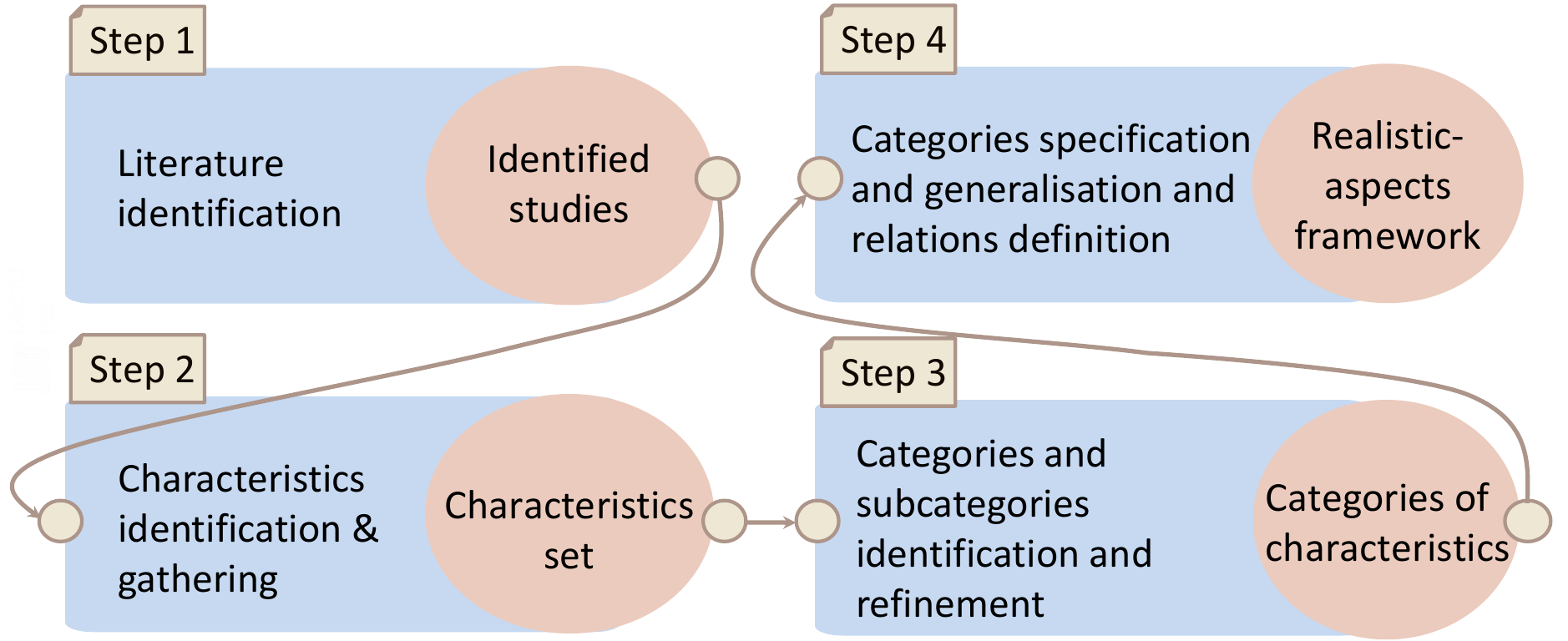}
	\caption{Our methodology for defining the realistic-aspects framework. Rectangles represent the steps, circles represent the output of each step, and arrows illustrate that the output of each step represents the input of the following one.}
	\label{fig:approach}
\end{figure}

 Our methodology is illustrated in Figure~\ref{fig:approach}. The first step is to identify the literature from which we can obtain initial ideas about realistic aspects of planning domains as discussed in the literature. We start with literature known to us and then use relevant terms to search for and explore other relevant studies. We are interested in studies that analyse the requirements and characteristics of application domains (e.g., building automation, smart homes, ubiquitous computing) by following a systematic way and/or developing a framework. Another type of research we are interested in focuses on characteristics that exist in real-world planning domains generally, i.e., not in a specific domain. A third strand of research that we explore focuses on providing quality measures to assess the quality of planning knowledge or that discuss what aspects can define the usefulness of the domains. The last strand of research that interests us contains some works that provide a systematic process of knowledge engineering and modelling of AI planning domains. The output of \textbf{Step 1} is 20 identified studies.

We use the set of identified studies to extract statements about realism in planning domains, and then identify and gather realistic aspects as described in these studies (\textbf{Step 2}). The output of this step is a collection of realistic characteristics. After that, we follow a descriptive research method to depict, describe, and organise the collected aspects~\cite{wazlawick2010reflections,glass1984descriptive}. Thus, in \textbf{Step 3}, we categorise the gathered aspects based on their relevance to one another. We also refine the gathered information of each category by combining similar aspects and distinguishing different concerns. In this step, each category is annotated with a common feature that describes all the aspects within the category. Additionally, in this step, we further extract subcategories as defined in the corresponding identified studies. The output of \textbf{Step 3} is a collection of identified categories and subcategories. Lastly, in \textbf{Step 4}, we identify the different relations between the categories, subcategories, and aspects. We organise these as a hierarchy of aspects and their categories with varying degrees of granularity. The final outcome of our approach is a conceptual framework represented by the hierarchy. We call this framework the \textit{realistic-aspects framework}. 

\subsection{Running Example: Sustainable Buildings Domain}
Sustainable buildings are smart buildings whose operation depends on the effectiveness and efficiency of their Building Management Systems (BMSs). These are computer-and-device-based control systems concerned with monitoring, storing, and communicating data, in addition to supervising, controlling, and automating the various systems in buildings~\cite{levermore2000building}. Examples of devices in sustainable buildings include sensors (e.g., position and temperature sensors), and actuators (e.g., switches on ceiling lamps). Systems in sustainable buildings can include Heating, Ventilation, Air conditioning (HVAC), lighting, access control, security, electrical, and other interrelated systems. The main objectives of BMSs include increasing safety, improving people's productivity, cutting energy consumption; hence preserving finite resources, using non-carbon sources when possible to lower the CO2 footprint, and lowering the costs of consumers and businesses while preserving users' comfort~\cite{georgievski2017planning,fiorini2019energy}. This is especially true for buildings that are connected to a smart grid, which makes it possible to include renewable sources and provides dynamic pricing and energy offers coming from competing providers~\cite{georgievski2015coordinating}. The advent of the Internet of Things (IoT) and advances in AI offer significant opportunities to improve the limited control capabilities offered by current building management systems, such as the reactive control and feedback mechanisms~\cite{georgievski2017planning,georgievski2023babtp}.

\section{The Framework} 
\label{sec:realistic-aspects}

We gather and extract realistic aspects of planning domains from the identified relevant studies. We categorise these realistic aspects into seven main categories based on their relevance to each other, namely: Objectives, Tasks, Quantities, Determinism, Agents, Constraints, and Qualities. We provide details on identified aspects and categories per study in Appendix~\ref{apx:table}.\footnote{Note that the categories and subcategories included in the table are the ones identified directly from the studies before any refinement and/or generalisation.}

\begin{figure}
	\centering
        \includegraphics[width=\textwidth]{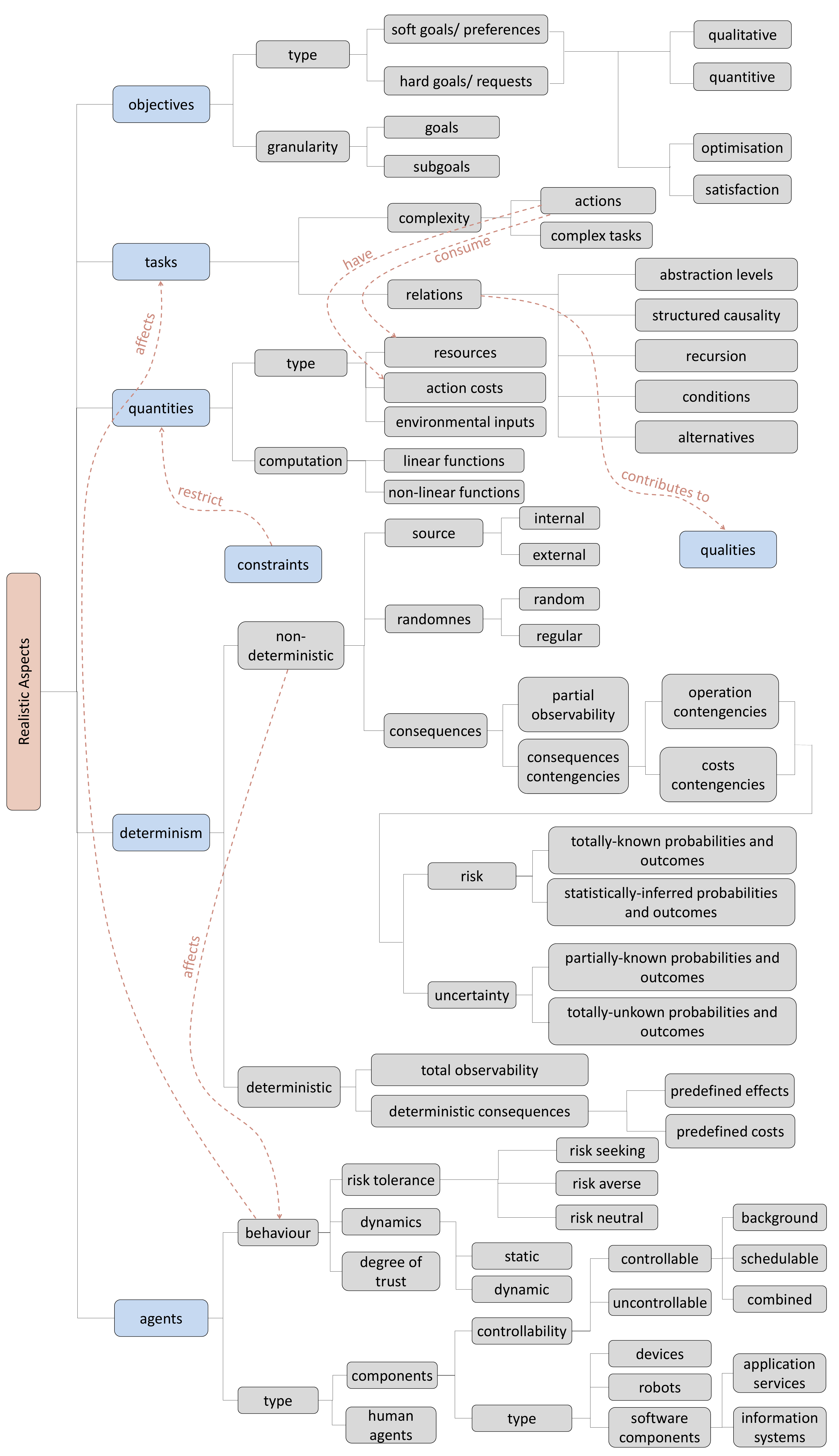}
	\caption{Realistic-Aspects Framework (Part One). Rectangles represent the aspects, blue rectangles are the aspects of the highest level, and dashed arrows define the relationships between the aspects.}
	\label{fig:aspects1}
\end{figure}

\begin{figure}
	\centering
        \includegraphics[width=\textwidth]{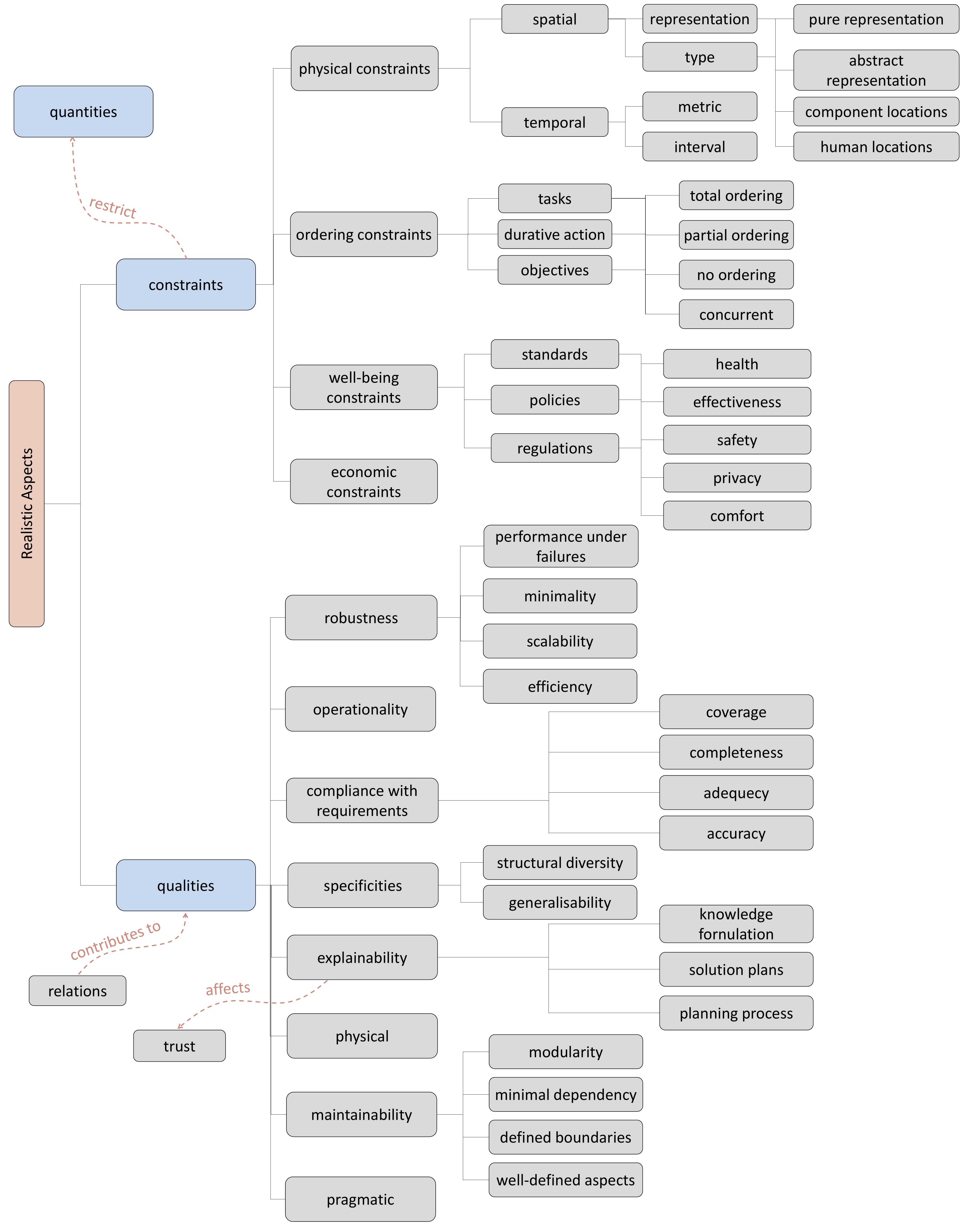}
	\caption{Realistic-Aspects Framework (Part Two).}
	\label{fig:aspects2}
\end{figure}

We use the identified aspects and categories to develop a conceptual framework in the form of a hierarchy of realistic aspects for planning domains. Figures~\ref{fig:aspects1} and~\ref{fig:aspects2} show the realistic-aspects framework. We split the hierarchy into two figures for better readability. The categories that we identified in \textbf{Step 3} form the highest level of the hierarchy. In the following, we organise the discussion of the realistic-aspects framework per category at the highest level.

\subsection{Objectives}
The first high-level category is related to the \textit{Objectives} of planning domains. We categorise the Objectives based on their \textit{Types} and \textit{Granularity}. For the first category, we distinguish two types of goals as suggested in~\cite{georgievski2016automated}. The first is \textit{Soft Goals} or \textit{Preferences}, which represent the non-mandatory user's desires that should be considered when solving planning problems in the domain. Considering our running example domain, the BMS might have the soft goal of keeping the offices in a building clean. The second is \textit{Hard goals} or \textit{Requests}, which, unlike the Preferences, define a mandatory behaviour of the planning system according to which planning problems in the domain should be solved. A hard goal in the Sustainable Buildings domain could be to maintain the energy consumption in the building under a certain threshold. 

We further classify the goals into \textit{Qualitative} and \textit{Quantitative} goals. That is, in some domains, it might be required to express goals in terms of qualities that the system should meet, such as improving the comfort of the building occupants, or in terms of exact quantities that should be maintained or reached, such as minimising the operation costs of the building to a certain value per year. Additionally, goals, whether they are soft or hard, can be of an \textit{Optimisation} nature or \textit{Satisfaction} nature. In the first case, the goal is to find an optimal solution to the planning problem. For example, in the Sustainable Buildings domain, according to the market prices, the amount of stored energy, and the weather forecast, the goal could be to find an optimal plan that minimises the energy consumption of the building. In the second case, the goal is to satisfy the domain goal to a certain degree. For example, given the aforementioned building conditions, the goal could be to compute plans that do not exceed a certain threshold of energy consumption.

For Granularity, in some application domains, we might need to express \textit{Subgoals} of bigger goals. For example, the goal of maintaining cleanness in the offices could be a subgoal of a more coarse-grained goal of increasing occupants' comfort. The latter can have another subgoal, such as the quantitative subgoal of maintaining the office temperature between 15-20$^{\circ}$.

\subsection{Tasks} \label{subsec:tasks}
The second high-level category of the hierarchy is related to the \textit{Tasks} performed in the application domain. We categorise Tasks into two subcategories based on their \textit{Complexity} and the \textit{Properties of Relations} between them, respectively. In the Complexity subcategory, we distinguish between \textit{Actions} and \textit{Complex Tasks}. Actions are the simplest kind of tasks that can be performed directly, such as pulling up the blinds. Complex Tasks, on the other hand, are more complex than Actions and cannot be performed directly, but are needed as they represent domain-specific knowledge. The Complex Tasks eventually depend on the tasks that can be performed directly. For example, in the Sustainable Buildings domain, the task of conducting a meeting in a meeting room is a complex task. It can be realised by performing the tasks of adjusting the light intensity in the room, adjusting the temperature, and turning on the projector.

Unlike in the Complexity subcategory, aspects in the Properties of Relations subcategory are interrelated and not mutually exclusive. We distinguish five relation properties, which are \textit{Abstraction Levels}, \textit{Structured Causality}, \textit{Recursion}, \textit{Alternatives}, and \textit{Conditions}~\cite{georgievski2017planning,alnazer2022bringing}. All these properties are related to the idea of achieving complex tasks by performing smaller tasks, i.e., the refinement of complex tasks into subtasks, which can be Complex Tasks and/or Actions. In particular, these properties characterise the relations between Complex Tasks and their subtasks. The Abstraction Levels property indicates that complex tasks represent an abstraction level to the subtasks they can be refined to. Structured Causality is a by-product of the Abstraction Levels property since expressing tasks knowledge in different abstraction levels leads to having causal reasoning between the different tasks. Recursion defines the relation between a Complex Task and itself. That is, it defines that some Complex Tasks can be performed by recursively refining them until reaching a certain condition. For example, the complex task of increasing the light intensity in the meeting room when we have multiple lamps can be refined into the action of turning on a lamp and the complex task of increasing the light intensity again. The recursion stops, for example, when reaching a certain light intensity or when having no more lamps to light in the room. The Alternatives property means that there might be multiple ways in which the same Complex Task can be achieved. For example, the complex task of increasing the light intensity can be performed by refining it to the task of pulling up the blinds or by refining it to the alternative task of turning on lamps. Finally, the Conditions property represents the situation where refining a Complex Task in a certain way has conditions that should be satisfied. For example, turning on lamps to increase the light intensity can only be done if there is not enough light coming from windows, i.e., if pulling up the blinds does not increase the light intensity to the required intensity level.

\subsection{Quantities}
The third high-level category of the hierarchy is related to \textit{Quantities}. We categorise Quantities in planning domains based on their \textit{Type} and how they should be computed. We call the latter subcategory \textit{Computation}. In the first case, we distinguish three types of quantities; \textit{Resources}, \textit{Action costs}, and \textit{Environmental Inputs}. Resources are quantities that define a bound to the allowed or possible consumption. They can be, for instance, money, fuel, energy, and time. In the Sustainable Buildings domain, the budget dedicated to the building operation represents a resource. Action costs represent the resource consumption incurred by performing actions. For example, if we consider the energy that is stored locally in the building as a resource, the action of turning on the heating system using locally stored energy results in a cost that equals the amount of consumed energy. The Environmental Inputs represent all measurable characteristics of the planning domain environment and the tasks performed in the domain. Examples of environmental inputs in the Sustainable Buildings domain include indoor temperature, CO$_2$ level, humidity, light intensity, energy demand, and battery capacity~\cite{georgievski2023babtp}. Similarly, drivers' trust in the autonomous vehicle is considered an environmental input~\cite{alnazer2022role}. We compute quantities based on the real-world concepts they express, where either \textit{Linear} or \textit{Non-linear functions} should be used~\cite{georgievski2023babtp}. For example, calculating temperature, battery charging, and tariff change requires using non-linear functions.

\subsection{Determinism}
The fourth high-level category of the hierarchy is related to whether the planning domain is deterministic or not. When the application domain is \textit{Deterministic}, all conditions of the environment are \textit{Totally observable} at all times. Additionally, in deterministic environments, actions work exactly as expected, i.e., they have \textit{Predefined Consequences} and \textit{Predefined Costs}. For example, in the Sustainable Buildings domain, if the heating system is 100\% reliable, the action of turning it on will always lead to the heating system being turned on. Furthermore, if the energy prices are known with 100\% certainty, the operation cost of the heating system, i.e., the cost of the action is predefined with certainty.

Most real-world planning domains are, however, \textit{Non-deterministic}~\cite{alnazer2022risk}. We categorise the non-determinism of planning domains based on the \textit{Source}, \textit{Randomness}, and \textit{Consequences} of non-determinism. Sources might be \textit{Internal}, meaning, they are related to the performer of the actions. These include, for example, internal malfunctions, and unreliability or limited capabilities of the agents. Sources might otherwise be \textit{External}, i.e., related to environmental conditions that are external to the agent, such as the non-determinism of the weather conditions. The Randomness subcategory defines whether the non-determinism source is \textit{Regular}, i.e., changes all the time, such as weather conditions, energy demand and market prices~\cite{fiorini2019energy,fiorini2020predictive}, or it is totally \textit{Random}, such as a malfunction in the battery storage. The Consequences category defines the effects that the non-determinism sources have on the planning domain. The first type of consequences is related to \textit{Partial Observability} of the surrounding environment. For example, an internal malfunction in the sensors responsible for detecting whether the person is working on his/her PC can lead to partial knowledge about the current conditions of the environment. The second type of Consequences is related to \textit{Action Contingencies}, i.e., actions not working as expected~\cite{kaldeli2013coordinating}. These can be \textit{Effect Contingencies} and/or \textit{Cost Contingencies}, which means that actions do not have predefined effects on the environment and do not consume resources as expected, respectively. For example, turning on the radiator to heat the office might lead to the radiator not being turned on due to internal malfunction. Additionally, turning on the radiator might have costs that are hard to predefine with certainty. The reason for this could be not having enough locally stored energy due to unexpected weather conditions. This necessitates buying energy, which incurs costs that depend on the market prices.

With the existence of non-determinism in the environment, we can have a full spectrum of the degree of knowledge that is available about the \textit{Action Consequences}. We might have \textit{Full Knowledge} about the probability distribution and outcomes of actions or we might be able to \textit{Statically infer} them. In both cases, we have \textit{Risk} involved in the domain. We might, however, have only \textit{Partial} or even \textit{No Knowledge} about the probability distribution of the action outcomes. In these cases, we have \textit{Uncertainty} in the domain. For the distinction between \textit{Risk} and \textit{Uncertainty} in AI planning, see~\cite{alnazer2022risk}.

\subsection{Agents} \label{subsec:agents}
The fifth high-level category is related to \textit{Agents}, which are the performers of actions. We classify agents based on their \textit{Type} and \textit{Behaviour}. For the Type, we distinguish between \textit{Human Agents} (e.g., occupants of the building) and non-human agents, which we refer to as \textit{Components}. Components are further categorised based on their \textit{Type} and \textit{Controllability}. Types of components include \textit{Devices} (e.g., actuators and batteries to store energy), \textit{Robots} (e.g., teleconferencing and cleaning robots), or \textit{Software Components}. The latter represents either \textit{Application Services} that can be commercial (e.g., Microsoft PowerPoint) or have a specific purpose (e.g., application services to control the desired thresholds of lighting and heating), or \textit{Information Systems} (e.g., platforms used to analyse smart building data and display them in a dashboard for monitoring)~\cite{georgievski2016automated}. 

Regarding Controllability, Components in planning domains can be either \textit{Controllable} or \textit{Uncontrollable}~\cite{georgievski2023babtp}. Controllable components can be controlled directly and are mostly internal to the application domain. These include components that run in the \textit{Background} all the time (e.g., water heating in buildings), \textit{Schedulable Components} (e.g., dishwashers), or \textit{Combined} components that can run in the background and be scheduled (e.g., space heating in buildings). On the other hand, Uncontrollable components are components whose operations cannot be controlled directly but depend mainly on conditions external to the application domain. An example of uncontrollable components in smart buildings is solar panels whose operations depend on weather conditions.

When dealing with domains that involve risk and uncertainty, the domain knowledge should reflect the \textit{Behaviour} of the agents. That is, when risk and uncertainty exist in the domain, agents can have different preferences on how to make decisions during planning. These preferences are related to the agents' \textit{Risk Tolerance} and the \textit{Degree of Trust} they have in the planning system to make the right choices. We classify the agent's risk tolerance based on its \textit{Degree} and \textit{Dynamics}. For the Degree, we distinguish three risk tolerance degrees, namely \textit{Risk-seeking}, \textit{Risk-averse}, and \textit{Risk-neutral}. For example, let us assume that a risk-seeking agent is confronted with two choices that have the same expected value of the outcomes. The first choice has a 100\% probability of its outcome, i.e., it has a guaranteed outcome. However, although the second choice has the same expected value of outcomes, the probability of having a good outcome is very low. In this case, the risk-seeking agent will take risks and choose the second option hoping to end up with a good outcome. On the other hand, a risk-averse agent will avoid taking risks and prefer the first choice as it is guaranteed. A risk-neutral agent will only consider the expected value of the outcomes and thus will be indifferent to the risk involved in each option, i.e., it will be indifferent to the two options. Considering our running example, the building can get its electricity from two different sources; either from stored electricity generated locally by renewable resources or from electric utilities with varying prices and energy offers. The choice of whether to consume locally generated electricity or to purchase it from outside sources should be made under uncertainty about the future market prices and the future weather forecast. Let us assume that the building management system has the following information: there is a high probability of having a cloudy next day, i.e., no energy can be stored in the batteries and a small probability of having a sunny day. However, there is information about the day-ahead prices offered by different providers. The building management system might follow a risk-seeking attitude and decides not to purchase energy and rely on the small probability of having a sunny day, i.e., using solar energy.

Regarding the Dynamics of the agent's risk tolerance, the agent might have \textit{Static} or \textit{Dynamic} risk tolerance. Static risk tolerance means that the agent will have the same degree of risk tolerance in one-shot planning with the same domain conditions. On the other hand, Dynamic risk tolerance changes during the one-shot planning based on some factors, such as the resource amount remaining in the domain. For example, the BMS might be making all choices during planning based on risk-seeking tolerance, but once the locally stored energy goes under a certain threshold, it will become less risk tolerant.

\subsection{Constrains}
The sixth high-level category is about the \textit{Constraints} in the application domain. We distinguish four different classes of constraints: \textit{Physical Constraints}, \textit{Ordering Constraints}, \textit{Well-being Constraints}, and \textit{Economic Constraints}. Physical Constraints relate agents and the actions they perform to each other with respect to space (i.e., \textit{Spatial Constraints}) and time (i.e., \textit{Temporal Constraints}). For example, a spatial constraint on the action of an occupant moving from the corridor to the office defines that the agent should be in the corridor before moving and will be in the office after performing the action. Another example where the agent is a component is a spatial constraint on the actions of the actuator that controls the blinds. That is, for the actuator to open the blinds, the actuator should be attached to the blinds. Spatial Constraints can be represented either abstractly or purely~\cite{aiello2007spatial}. An abstract representation is a representation without considering any geometrical or physical laws, where actions are considered to be, for example, instantaneous. An example of this is assuming that an occupant's movement from the corridor to the office or the battery storing the energy generated from solar panels happens instantaneously. On the other hand, a pure representation considers the geometrical or physical laws and sometimes the spatiotemporal properties~\cite{andreka2007logic}. This means that representing the movement of an occupant requires considering the time needed for the movement and the spatial arrangement of the building, i.e., the locations of the corridor and the office.
According to this, we categorise the Spatial Constraints based on the agents performing the spatially constrained actions into \textit{Human Locations} and \textit{Component Locations}. We also categorise them based on their representation into \textit{Abstract Representation} and \textit{Pure Representation}. 

Similar to Spatial Constraints, Temporal Constraints might be defined with respect to the actions executed by agents and the relations between these actions. We distinguish two categories of Temporal Constraints \textit{Metric Constraints} and \textit{Interval Constraints}. The first type defines an absolute time point at which an action should be performed. For example, actions that turn off the lights in the whole building should be performed at the end of the working day. This type of constraints can also restrict the relations between different actions with respect to absolute time points. For example, we might have a constraint that restricts the action of opening the blinds in the room and turning on the room lights at the same time since this might lead to a waste of energy. Interval Constraints restrict the start and end time of actions. We refer to these actions as \textit{Durative Actions}. In other words, they are defined on the time interval in which actions are performed. For example, there might be a temporal constraint that restricts when the cleaning robot can start and end cleaning in the office to avoid disturbing office occupants. Additionally, Temporal Constraints can restrict the relations between durative actions with respect to time. For example, there might exist a temporal constraint that requires the start of the air conditioner operation using locally stored energy to follow the end of the storing energy in batteries operation. Spatial and Temporal Constraints can also restrict the existence of agents with one another even if these agents are not performing actions. For example, in a health-sensitive situation, regulations might be in place that restrict the existence of more than a certain number of people in the same location (spatial constraint) at the same time (temporal constraint).

The second class of constraints is related to the \textit{Ordering} between Objectives, Complex Tasks, or Durative Actions. These can have a \textit{Strict Total Ordering}, \textit{Partial Ordering}, \textit{Require Concurrency}, or can be \textit{Unordered}. Ordering between objectives can represent the importance of these objectives. For example, the objective of maintaining occupants' safety cannot be compromised by satisfying other objectives, such as improving the occupants' comfort. In the context of objectives, concurrency ordering means that objectives have the same exact importance. For Durative Actions, we can have, for example, a strict ordering between the operation of storing energy and the operation of operating the air conditioner, or we can have a flexible order, i.e., unordered constraints, as long as the battery has enough energy. We can also have a partial order between durative actions, where we define a certain degree of order among them, i.e., some actions have a specific order of execution while others do not. This also applies to complex tasks.

The third class of constraints is called \textit{Well-being Constraints}. These constraints are derived from regulations, policies, or standards of the application domain and aim at improving users' comfort, privacy, health, safety, and effectiveness~\cite{georgievski2017planning}. For example, turning lights on/off and controlling their intensity in the office can follow some standards for lighting in indoor workplaces~\cite{georgievski2017planning}. 
The fourth class of constraints is defined on the actions performed in the application domain and is of economical nature. Thus, we call it \textit{Economical Constraints}. An example of these kinds of constraints can be a constraint on the maximum energy consumption, i.e., cost induced as a result of performing a specific operation.
 
The classes of constraints are not mutually exclusive. Well-being constraints, for example, can restrict the existence of multiple people in the same location, the existence of certain components with human beings at the same location, or the existence of people and/or components at the same time~\cite{georgievski2023babtp}.

\subsection{Qualities}
The last high-level category of the framework is concerned with the \textit{Qualities} of the different aspects of a planning domain and the overall planning system. Identifying the required qualities of the planning system, including all its elements, contributes directly to our goal of providing support to knowledge engineers and software engineers in the process of designing and developing the planning system.

We categorise the Qualities into eight classes. The first class is the \textit{Robustness} class, which reflects how robust planning is to changes. This includes the domain's (1) \textit{Minimality}, (2) \textit{Scalability}, (3) \textit{Efficiency}, and (4) \textit{Performance under Failures}. These define (1) how compact is the model of the domain, such that the domain knowledge is modelled efficiently and there are no additional or unnecessary constructs; (2) how well the planning system enables solving planning problems of increased complexity; (3) how efficiently planning problems in this domain can be solved; and (4) how robust the planning system is against failures in some parts of it, respectively. These aspects characterise the planning system as a whole, but at the same time, are affected by the planning domain itself. For example, the efficiency quality depends partially on the planning system, but the domain structure, minimality, and encoding language may have a great impact on the efficiency of the planning system.

The second class is related to the compliance of the planning system and its elements with the requirements of the application domain. We call this class \textit{Compliance with Requirements} and it includes: (1) \textit{Coverage}, (2) \textit{Completeness}, (3) \textit{Accuracy}, and (4) \textit{Adequacy}. Coverage defines to what extent the planning domain model covers all required aspects of the application area. Completeness means that the domain model enables the generation of all (and only) solution plans that are correct with respect to the domain specifications. Accuracy ensures that the domain model is a valid representation of the domain specifications, meaning it encodes all the aspects that are correct and relevant for the application domain. Adequacy is related to the expressive power of the planning modelling language to represent the requirements within a planning domain model in sufficient detail so that a complete planning domain model can be expressed. For more details and formal definitions of Completeness, Accuracy, and Adequacy aspects, see~\cite{mccluskey2017engineering,silva2020formal}.

The third class is related to the \textit{Specificities} of the application domain. This class includes \textit{Structural diversity} and \textit{Generalisability}. These are related to how well the planning domain model reflects the specificities of the application domain~\cite{alnazer2022bringing} and how easy it is to be generalised to enable solving planning problems with fewer or more specificities, respectively. For example, in the Sustainable Building domain, specificities of the domain include the task of preparing the meeting room. This task is specific to this domain. Generalisability in this domain can be achieved, for example, by having different abstraction levels of tasks (see Section~\ref{subsec:tasks}).

The fourth class is related to the \textit{Maintainability} of the planning system in general and the planning domain in particular. This class defines how easily the planning domain model can be modified to handle new requirements and includes (1) \textit{Modularity}, (2) \textit{Scope/Defined Boundaries}, (3) \textit{Well-defined Aspects}, and (4) \textit{Minimal Dependency}. Modularity is achieved when the planning system/domain model has clearly divided modules, thus, it is easier to modify a specific module without the need to change the whole planning system/domain model. This is also related to the planning system/planning domain model being scoped, i.e., having well-defined boundaries and aspects that precisely reflect the specific requirements of the application domain. Lastly, the fewer dependencies between the different modules in the planning system/domain model, the easier it is to be maintained. 

The fifth class is related to the \textit{Explainability} of the planning system behaviour and the planning knowledge, especially to non-experts. This is related to the (1) formation of the domain knowledge, (2) the formation of the plans computed in the domain, and (3) the planning process itself~\cite{alnazer2022role}. For example, the structured causality between tasks in the domain enables causal reasoning, which makes it easy to track and explain the behaviour of the system.

The last three classes are related to \textit{Physical}, \textit{Pragmatic}~\cite{vallati2021quality}, and \textit{Operationality} aspects~\cite{mccluskey2017engineering,vallati2021quality}. The Physical class focuses on maximising the availability and accessibility of the planning domain models to interpreters to make sense of and revisit domain models. This class aims at maximising the availability of planning knowledge models such that future validations and evaluations can be performed. The Pragmatic class is related to the analysis of domain models after the design phase and focuses on how the changes made to domain models, as a result of using them in planning systems, can lead to the discovery of missing requirements. Operationality is related to the ability of planning systems to reason upon domain models and generate solution plans using bounded computational resources (e.g., memory usage). It focuses mainly on the quality and shape of the resulting plans and the speed of plan generation. For more details and formal definitions of the Physical, Pragmatic, and Operationality aspects, see~\cite{mccluskey2017engineering,vallati2021quality}.

\section{Conclusions and Future Work}\label{sec:conclusions}
Despite the research advancements in AI planning, applying AI planning systems to solve real-world planning problems seems to be still challenging. Improving this situation requires understanding and considering the aspects that characterise real-world planning domains in all development phases of planning systems. In particular, knowledge engineers and software engineers should be supported in making informed choices when reasoning about relevant and realistic aspects of application domains that must be considered when developing planning systems. Currently, this process is not supported. The main obstacles seem to be the broad range of aspects of planning domains and the lack of unified notions of what makes a planning domain realistic.

We took a step forward in this direction by introducing a framework that conceptualises the notion of planning domains' realism. We gathered and analysed information about planning domains from existing literature and, consequently, developed a framework that categorises a large number of realistic aspects in multi-level granularity. The framework provides a common notion of planning domains' realism, highlights some aspects (e.g., uncertainty and risk) that are simplified and/or neglected in the literature of AI planning, can drive the development of service-oriented AI planning systems, and offers means for comparing different planning systems. For future work, we plan to synthesise metrics that can quantitatively evaluate the realism of planning domains and empirically verify and validate the conceptual framework on realistic application domains. We also want to explore the framework's usability for the design of service-oriented AI planning systems.
\appendix
\setcounter{secnumdepth}{0}
\section{Appendix}\label{apx:table}
Table~\ref{table:realistic-aspects1} shows the identified studies, identified aspects in each study, subcategories as defined in the identified studies, and the category to which each aspect and subcategory belongs, respectively.

\begin{table}[]
\caption{Realistic aspects of planning domains in literature with the identified categories and subcategories.}
\label{table:realistic-aspects1}
\begin{tabular}{|P{1,1cm}|P{6,2cm}|P{1,9cm}|P{2,1cm}|}
\hline
Studies & Characteristics & \multicolumn{1}{l|}{\;\;Subcategories\;} & Category \\ \hline
\multirow{12}{*}{\cite{georgievski2016automated}} & preferences & \multirow{2}{*}{-} & \multirow{2}{*}{\begin{tabular}[c]{@{}l@{}}behavioural\\ inputs\end{tabular}} \\ \cline{2-2}
 & Requests &  &  \\ \cline{2-4} 
 & device operations & \multirow{5}{*}{-} & \multirow{5}{*}{\begin{tabular}[c]{@{}l@{}}behavioural \\ outputs\end{tabular}} \\ \cline{2-2}
 & human operations &  &  \\ \cline{2-2}
 & robot operations &  &  \\ \cline{2-2}
 & application operations &  &  \\ \cline{2-2}
 & information operations &  &  \\ \cline{2-4} 
 & (s) object and human locations & \multicolumn{1}{c|}{\multirow{2}{*}{\begin{tabular}[c]{@{}c@{}}(s)patial, \\ (t)emporal\end{tabular}}} & \multirow{2}{*}{\begin{tabular}[c]{@{}l@{}}physical\\ properties\end{tabular}} \\ \cline{2-2}
 & \begin{tabular}[c]{@{}c@{}}(t) time points (metric constraints) \\ and qualitative relations (intervals)\end{tabular} & \multicolumn{1}{l|}{} &  \\ \cline{2-4} 
 & unexpected events & \multirow{3}{*}{-} & \multirow{3}{*}{Uncertainty} \\ \cline{2-2}
 & partial observability &  &  \\ \cline{2-2}
 & operations contingencies &  &  \\ \hline
 \multirow{3}{*}{\cite{silva2020formal}} & definition of goals and subgoals & - & \multicolumn{1}{c|}{-} \\ \cline{2-4} 
 & accuracy & - & \multicolumn{1}{c|}{-} \\ \cline{2-4} 
 & adequacy & - & \multicolumn{1}{c|}{-} \\ \hline
\multirow{2}{*}{\cite{kaldeli2009extended,kaldeli2013coordinating}} & definition of goals and subgoals & - & \multicolumn{1}{c|}{-} \\ \cline{2-4} 
 & expressing temporal constructs in goals & - & \multicolumn{1}{c|}{-} \\ \hline
 
%

\cite{hoffmann2006engineering} & structural diversity & - & \multicolumn{1}{c|}{-} \\ \hline
\multirow{2}{*}{\cite{mccluskey2004knowledge}} & objects, relations, properties, and constraints & - & static knowledge \\ \cline{2-4} 
 & object behaviour & - & dynamic knowledge \\ \hline
\multirow{3}{*}{\cite{alnazer2022risk}} & risk & - & \multicolumn{1}{c|}{-} \\ \cline{2-4} 
 & uncertainty & - & \multicolumn{1}{c|}{-} \\ \cline{2-4} 
 & risk attitude & - & \multicolumn{1}{c|}{-} \\ \hline
\cite{alnazer2022role} & trust & - & \multicolumn{1}{c|}{-} \\ \hline
\multirow{3}{*}{\cite{alnazer2022bringing}} & structural diversity & - & \multicolumn{1}{c|}{-} \\ \cline{2-4} 
 & action costs & - & \multicolumn{1}{c|}{-} \\ \cline{2-4} 
 & alternatives & - & \multicolumn{1}{c|}{-} \\ \hline
\cite{vallati2021defence} & explainability & - & \multicolumn{1}{c|}{-} \\ \hline
\cite{chrpa2020modeling} & efficiency & - & \multicolumn{1}{c|}{-} \\ \hline
\cite{gregory2016domain} & action costs & - & \multicolumn{1}{c|}{-} \\ \hline
\multirow{12}{*}{\cite{georgievski2023babtp}} & ambient operations & \centering \multirow{3}{*}{-} & \multirow{3}{*}{\begin{tabular}[c]{@{}l@{}}building \\ operations\end{tabular}} \\ \cline{2-2}
 & electrical equipment operations &  &  \\ \cline{2-2}
 & green operations &  &  \\ \cline{2-4} 
 & (c) background, schedulable, and combined & \multicolumn{1}{l|}{\begin{tabular}[c]{@{}l@{}}(c)ontrollable, \\ (u)ncontrollable\end{tabular}} & \begin{tabular}[c]{@{}l@{}}building\\ properties\end{tabular} \\ \cline{2-4} 
 & linear & \multirow{2}{*}{-} & \multirow{2}{*}{quantities} \\ \cline{2-2}
 & non-linear &  &  \\ \cline{2-4} 
 & well-being constraints & \multirow{5}{*}{-} & \multirow{5}{*}{constraints} \\ \cline{2-2}
 & operation constraints &  &  \\ \cline{2-2}
 & temporal: required concurrency &  &  \\ \cline{2-2}
 & ordering constraints &  &  \\ \cline{2-2}
 & business constraints &  &  \\ \cline{2-4} 
 & strict-total ordering & \multicolumn{1}{c|}{-} & objectives \\ \hline
\end{tabular}
\end{table}

\begin{table}[]
\begin{tabular}{|P{1,1cm}|P{6,2cm}|P{1,9cm}|P{2,1cm}|}
\hline
\multirow{9}{*}{\cite{georgievski2017planning}} & occupant activity & \multicolumn{1}{c|}{-} & \multicolumn{1}{c|}{-} \\ \cline{2-4} 
 & building properties & \multicolumn{1}{c|}{-} & \multicolumn{1}{c|}{-} \\ \cline{2-4} 
 & activity area & \multicolumn{1}{c|}{-} & \multicolumn{1}{c|}{-} \\ \cline{2-4} 
 & building condition & \multicolumn{1}{c|}{-} & \multicolumn{1}{c|}{-} \\ \cline{2-4} 
 & social and economic (resource management) & \multicolumn{1}{c|}{-} & \begin{tabular}[c]{@{}l@{}}quality \\ conditions\end{tabular} \\ \cline{2-4} 
 & modularity/abstraction & \multicolumn{1}{c|}{-} & \multicolumn{1}{c|}{-} \\ \cline{2-4} 
 & structured causality & \multicolumn{1}{c|}{-} & \multicolumn{1}{c|}{-} \\ \cline{2-4} 
 & ordering control & \multicolumn{1}{c|}{-} & \multicolumn{1}{c|}{-} \\ \cline{2-4} 
 & recursion & \multicolumn{1}{c|}{-} & \multicolumn{1}{c|}{-} \\ \hline
\multirow{14}{*}{\cite{aiello2022software}} & explainability & \multirow{6}{*}{-} & \multirow{6}{*}{\begin{tabular}[c]{@{}l@{}}user-related\\ requirements\end{tabular}} \\ \cline{2-2}
 & improvement of sustainability &  &  \\ \cline{2-2}
 & safety &  &  \\ \cline{2-2}
 & privacy &  &  \\ \cline{2-2}
 & keeping the effectiveness of people &  &  \\ \cline{2-2}
 & comfort &  &  \\ \cline{2-4} 
 & ordering constraints & \multicolumn{1}{c|}{-}  & \multicolumn{1}{c|}{-} \\ \cline{2-4} 
 & \begin{tabular}[c]{@{}c@{}}business, administrative, user,\\ and system requirements\end{tabular} & \multicolumn{1}{c|}{-} & \multicolumn{1}{c|}{-} \\ \cline{2-4} 
 & \begin{tabular}[c]{@{}c@{}}minimize operation costs\\ and environmental impact,\\ and satisfy the power needs\end{tabular}  & \multicolumn{1}{c|}{-} & objectives \\ \cline{2-4} 
 & modularity & \multirow{5}{*}{-} & \multirow{5}{*}{\begin{tabular}[c]{@{}l@{}}non-functional \\ requirements\end{tabular}} \\ \cline{2-2}
 & well-defined components &  &  \\ \cline{2-2}
 & minimal dependencies &  &  \\ \cline{2-2}
 & increase workload &  &  \\ \cline{2-2}
 & perform under failures &  &  \\ \hline
\multirow{3}{*}{\cite{evans2014requirements}} & Express the degree of goal satisfaction & - & \multicolumn{1}{c|}{-} \\ \cline{2-4} 
 & domain boundaries (scope) & - & \multicolumn{1}{c|}{-} \\ \cline{2-4} 
 & flexibility, generality, and robustness & - & \multicolumn{1}{c|}{\;\;\;\;\;\;\;\;\;\;\;\;\;\;\;\;\;\;\;\;\;} \\ \hline
\multirow{4}{*}{\cite{chien1996real}} & understandable and modifiable plans & - & \multicolumn{1}{c|}{-} \\ \cline{2-4} 
 & temporal constraints: relative and absolute & - & \multicolumn{1}{c|}{-} \\ \cline{2-4} 
 & flexibility, generality, and robustness & - & \multicolumn{1}{c|}{-} \\ \cline{2-4} 
 & Express the degree of goal satisfaction & - & \multicolumn{1}{c|}{-} \\ \hline
\multirow{3}{*}{\cite{shah2013knowledge}} & efficiency/operationality & - & \multicolumn{1}{c|}{-} \\ \cline{2-4} 
 & maintenance and documentation & - & \multicolumn{1}{c|}{-} \\ \cline{2-4} 
 & clear and easy to understand by non-experts & - & \multicolumn{1}{c|}{-} \\ \hline
\multirow{2}{*}{\cite{vallati2021quality}} & encoding language & - & \multicolumn{1}{c|}{-} \\ \cline{2-4} 
 & \begin{tabular}[c]{@{}l@{}}semantic, syntactic, physical, pragmatic,\\  and operational qualities\end{tabular} & - & \multicolumn{1}{c|}{-} \\ \hline
\cite{mccluskey2017engineering} & \begin{tabular}[c]{@{}l@{}}accuracy, adequacy, completeness, \\ operationality, and consistency\end{tabular} & - & \multicolumn{1}{c|}{-} \\ \hline
\end{tabular}
\end{table}
\vspace{-0,15cm}

\bibliographystyle{splncs04}
\bibliography{bibliography}

\end{document}